 \documentclass[tablecaption=bottom,wcp]{jmlr} 

\usepackage{times}
\usepackage{latexsym}
\usepackage{float}
\usepackage{graphicx}
\usepackage{paralist}
\usepackage{caption}
\usepackage{subcaption}
\usepackage{url}
\usepackage{listings}
\usepackage{multicol}
\usepackage{todonotes}

\usepackage{booktabs}
\usepackage[load-configurations=version-1]{siunitx} 

\jmlryear{2019}
\jmlrsubmitted{April 6th, 2019}
\jmlrworkshop{LearnAut 2019} 

\title[Short Title]{Full Title of Article\titlebreak This Title Has
A Line Break}

\title{Automatic Inference of Minimalist Grammars using an SMT-Solver}

\author{\Name{Sagar Indurkhya\nametag{}} \Email{indurks@mit.edu}\\
 \addr MIT}

\begin{document}

\maketitle
\begin{abstract}
  We introduce (1) a novel parser for Minimalist Grammars (MG),
  encoded as a system of first-order logic formulae that may be
  evaluated using an SMT-solver, and (2) a novel procedure for
  inferring Minimalist Grammars using this parser. The input to this
  procedure is a sequence of sentences that have been annotated with
  syntactic relations such as semantic role labels (connecting
  arguments to predicates) and subject-verb agreement. The output of
  this procedure is a set of minimalist grammars, each of which is
  able to parse the sentences in the input sequence such that the
  parse for a sentence has the same syntactic relations as those
  specified in the annotation for that sentence. We applied this
  procedure to a set of sentences annotated with syntactic relations
  and evaluated the inferred grammars using cost functions inspired by
  the Minimum Description Length principle and the Subset principle.
  Inferred grammars that were \emph{optimal} with respect to certain
  combinations of these cost functions were found to align with
  contemporary theories of syntax.
\end{abstract}

\section{Introduction}
\label{sec:introduction}
Inspired by earlier formulations of grammars using logic
(\cite{Pereira:1983:PD:981311.981338, rayner1988using,
  stabler1993logical, rogers1998descriptive, graf2013local}) and
recent, substantive improvements in the performance of SMT-solvers
(\cite{de2011satisfiability, cadar2013symbolic}), we have developed a
novel procedure, with the form of a model of language acquisition
(\cite{chomsky1965aspects}), for automatically inferring Minimalist
Grammars (MG) (\cite{stabler1996derivational}) from a sequence of
sentences that have been annotated with syntactic relations for
predicate-argument structure and morphological agreement
.\footnote{Please contact the authors to obtain an implementation of
  the inference procedure introduced in this study.}
In this study, we report preliminary results that demonstrate our
inference procedures' capacity to acquire grammars that comport with
contemporary theories of minimalist syntax
(\cite{chomsky1995minimalist}).\footnote{For detailed presentations of
  minimalist syntax, see (\cite{adger2003core,
    hornstein2005understanding, radford2009introduction}).}

The remainder of this study is organized as follows: after reviewing
the MG formalism and prior work on modeling MGs with logic
(\sectionref{sec:minimalistgrammars}), we present our inference
procedure (\sectionref{sec:inferenceprocedure}) and use it to infer a
set of MG lexicons from the sequence of annotated sentences listed in
\tableref{table:input} (\sectionref{sec:experiment});
we identify members of the inferred set of MG lexicons that are
optimal with respect to cost functions inspired by the Minimum
Description Length (MDL) principle (\cite{grunwald2007minimum}) and
the Subset principle (\cite{berwick1985acquisition, wexler1993subset})
and present several examples of these optimal MG lexicons, \emph{one
  of which aligns with contemporary minimalist syntax, producing for
  each sentence in the input sequence a parse tree that matches
  standard syntactic analysis}.
Finally, in (\sectionref{sec:conclusion}) we discuss how our
procedure, which takes the form of a computational model of language
acquisition, may be applied to the evaluation of the Strong Minimalist
Thesis.\footnote{\cite{chomsky2008phases} The Strong Minimalist Thesis
  asserts that \emph{``language is an optimal solution to interface
    conditions that FL must satisfy; that is, language is an optimal
    way to link sound and meaning, where these notions are given a
    technical sense in terms of the interface systems that enter into
    the use and interpretation of expressions generated by an
    I-language.''} See also: \cite{chomsky2001derivation}.}

\section{Minimalist Grammars}
\label{sec:minimalistgrammars}

The Minimalist Grammar (MG) formalism, introduced in
\cite{stabler1996derivational}, is a well established formal model of
syntax inspired by \cite{chomsky1995minimalist}.
We chose to use this formalism because:
(i) MGs are mildly context-sensitive
(\cite{michaelis1998derivational}) and can model cross-serial
dependencies that arise in natural language
(\cite{Vijay-Shanker:1987:CSD:981175.981190, stabler2004varieties});
(ii) MGs can model displacement, a basic fact of natural language that
enables a phrase to be interpreted both in its final, surfaced
position, as well as other positions within a syntactic structure
(\cite{chomsky2013problems}).\footnote{A single phrase satisfying
  multiple interface conditions often requires that it undergo
  syntactic movement to establish a discontinuous structure (i.e. a
  chain) with multiple local relations; by the \emph{Principle of Last
    Resort}, movement is driven by morphological considerations -- e.g.
  morphological agreement (\cite{chomsky1995minimalist}).}

An MG consists of:
(i) a \emph{lexicon}, consisting of a finite set of atomic structures,
referred to as \emph{lexical items}, each of which pairs a phonetic
form\footnote{A phonetic form is either overt (e.g. a word in the
  sentence) or covert (i.e. unpronounced).} with a finite sequence of
(syntactic) features.\footnote{A feature has: (i) a value from a
  finite set of categories; (ii) a type, which is either
  \emph{selector}, \emph{selectee}, \emph{licensor} or
  \emph{licensee}, indicated by the prefix $=$, $\sim$, $+$ and $-$
  respectively; a $<$ or $>$ prefixed before a selector prefix
  indicates that the selector triggers left or right head-movement
  respectively. There is also a special feature, $C$, that serves to
  indicate the completion of a parse.}
(ii) \emph{merge}, a recursive structure building operation that
combines two structures to produce a new structure.\footnote{
  \emph{Merge} applies to two logically disjoint cases:
  (i) \emph{internal merge}, for the case in which one argument is a
  substructure of the other (i.e. they are not disjoint), requires
  that the consumed features for the two arguments be a
  \emph{licensor} and \emph{licensee}, with the former projecting.
  (ii) \emph{external merge}, for the case in which the two arguments
  are disjoint, requires that the consumed features for the two
  arguments be a \emph{selector} and \emph{selectee}, with the former
  projecting.}
Each application of \emph{merge} consumes the first feature from each
of its two arguments, requiring that the two features have the same
value; one of the two arguments then \emph{projects} its feature
sequence to the structure produced by merged.
To parse a sentence, a set of lexical items is drawn from the lexicon
and combined together, via the recursive application of \emph{merge},
into a single structure in which all of the features have been
consumed;
if the ordering of the phonetic forms in the resulting structure
aligns with the order of the words in the sentence being parsed,
then the structure is considered to be a valid parse of the
sentence.
See \figureref{fig:derivationA} and \figureref{fig:derivationB} for
examples of MG parses.

Finally, let us consider whether an MG may be modeled with
Satisfiability Modulo Theory (SMT) (\cite{barrett2018satisfiability}.
\cite{rogers1998descriptive} established that the axioms of GB theory
are expressible with Monadic Second Order logic (MSO); subsequently,
\cite{graf2013local} produced an MSO axiomatization for
MGs\footnote{\cite{graf2013local} also shows that constraints may be
  encoded in an MG lexicon if and only if they are MSO expressible.},
and notes that over finite domains these constraints may be expressed
with first order logic.
As this study only considers models with finite domains, we \emph{can}
develop a finite theory of MGs with an axiomatization based in part on
the MSO axiomatization of MGs developed by
\cite{graf2013local}.\footnote{Although we use the concept of
  \emph{slices}, presented in \cite{graf2013local}, our axiomatization
  does not utilize the first-order theory of finite trees (originally
  presented in \cite{backofen1995first}).}
We will express the theory with a multi-sort
quantifier-free\footnote{The axioms in the theory must be quantifier
  free as the SMT-solver cannot guarantee decidability for problems
  involving universal quantifiers; this is established via explicit
  quantification.} first-order logic extended with the theory of
uninterpreted functions (i.e. the theory of equality), allowing us to
model the theory with an SMT-solver and (decidably) identify
interpretations of models.

\section{Inference Procedure}
\label{sec:inferenceprocedure}
Our inference procedure takes the form of a computational model of
language acquisition (\cite{chomsky1965aspects}) consisting of:
%
(i) an initial state, $S_0$, consisting of a system of first-order
logical formulae that serve as axioms for deducing the class of
minimalist lexicons;
(ii) the input, consisting of a sequence of $n$ sentences, denoted
$I_1, I_2, \ldots, I_n$, each of which is annotated with syntactic
relations between pairs of words in the sentence;
(iii) a function, $Q$, that takes as input a state, $S_i$, and an
annotated sentence, ${I}_i$, and outputs the successor state,
$S_{i+1}$;
(iv) a function, $R$, that maps a state $S_i$ to a set of MG lexicons,
$G_i$, with the property that for each sentence $I_j$ in the input
sequence, each lexicon $L \in G_i$ can produce a parse $p_{j}^{L}$ such
that the syntactic relations in $p_{j}^{L}$ parse match those
specified in the annotation of $s_j$.\footnote{In the case of the
  initial state, $S_0$, since there are no constraints yet imposed by
  the input, $R(S_0)$ will map to the set of all minimalist lexicons.}
The procedure consumes the input sequence one annotated sentence at a
time, using $Q$ to drive the initial state, $S_0$, to the final state,
$S_n$; the function $R$ is then applied to $S_n$ to produce a set of
MG lexicons, $G_n$, that constitutes the output of the inference
procedure. (See Table-\ref{table:input} for an example of
input to the procedure)

We implemented this inference procedure by encoding an MG parser as a
system of first-order, quantifier-free logical formulas that can be
solved with an SMT-solver.\footnote{All logical formulas in this
  study, being used to encode finite models over bounded domains, are
  first-order and quantifier-free; this has the benefit that these
  formulas are decidable.}
This system of formulas is composed of formulas for MG parse trees
(see \sectionref{subsec:parsetreemodel}) that are connected (by way of
shared symbols) to a formula for an MG lexicon (i.e. $S_0$);
by imposing constraints on the formulas for parse trees, the set of
solutions to the lexicon formula is restricted.
Let us now review the role $Q$ and $R$ play in this.

When the inference procedure consumes an annotated sentence from the
input sequence, the function $Q$:
(1) instantiates a formula for a MG parse;
(2) translates the annotations for the sentence into (logic) formulas
that constrain the parse tree -- e.g. predicate-argument relations and
morphological agreement are translated into locality
constraints\footnote{The \emph{principle of syntactic locality}
  asserts that syntactic relations are established locally by merge
  (\cite{sportiche2013introduction}).}, and each sentence is
marked as declarative or interrogative, indicating which of two
pre-specified covert phonetic forms, $\epsilon_{CDecl}$ or
$\epsilon_{CIntr}$, must appear in a parse of the sentence (see Fig.
\ref{fig:derivationB} for an example);
(3) adds these new formulas to the existing system of formulas in
$S_i$ to produce $S_{i+1}$.

In order to compute the set of lexicons, $G_i = R(S_i)$, we used the
Z3 SMT-solver to solve for the set of lexicons satisfying the formulae
in $S_i$.\footnote{Z3 is a high-performance solver for Satisfiability
  Modulo Theories (SMT) that can solve first-order quantifier-free
  multi-sort logic formulas that may combine symbols from a set of
  additional logics defined by a number of background theories such as
  empty theory (i.e. the theory of uninterpreted functions with
  equality). See \cite{DeMoura:2008:ZES:1792734.1792766} for further
  reference.} Note that the inferred set $G_i$ is not enumerated;
rather, it exists implicitly in the model produced by the SMT-solver
(i.e. the solution to the system of logical formulas), and members of
this set may be filtered, searched and sampled by querying this model
using Z3; since the number of inferred lexicons is often exponentially
large due to symmetries, we do not enumerate the entire set of
lexicons; instead we use Z3 to sample lexicons from $G_i$.


\subsection{Modeling an MG Parse Tree}
\label{subsec:parsetreemodel}
We now provide an overview of a finite model of a minimalist parse
tree, based closely on the MG formalism, that we have developed using
a multi-sort first-order quantifier-free logic extended with the
theory of equality and uninterpreted functions.
The model consists of several sorts, uninterpreted functions acting
over these sorts, and a set of axioms constraining these functions
that every minimalist parse tree must satisfy;
additionally, the syntactic relations that annotate a sentence can
also be expressed as a set of axioms (first order logic formulas) that
further constrain the model.
An interpretation of the model (consisting of interpretations of the
uninterpreted functions) is thus a minimalist parse tree that accords
with the specified syntactic relations for a given sentence.
  
A (minimalist) parse tree\footnote{This tree corresponds to the
  derivation tree in the MG formalism.} is modeled as a labeled
directed acyclic graph\footnote{Each node in this graph corresponds to
  a node in the parse tree and the graph is constrained so as to have
  a single element with no out-going edges (which corresponds to the
  root of the parse tree); nodes with no incoming edges correspond to
  atomic syntactic structures.}, which is in turn modeled via (i) a
finite sort, members of which are nodes in the graph and (ii) a set of
(unary and binary) uninterpreted functions and predicates (acting over
the sorts), that establish labeled edges in the graph.\footnote{E.g.
  an uninterpreted binary predicate models dominance relations between
  nodes in the graph, and the transitive closure of this predicate
  establishes a binary tree in accordance with the Binary Branching
  Hypothesis (\cite{radford2009introduction}).}
Interpretations of these functions and predicates are constrained by a
set of axioms that include both: (a) an
axiomatization for the MG formalism; (b) axioms that aid in
expressing constraints imposed by interface conditions -- e.g. axioms
for structural configurations for predicate-argument structure and
projection of categories.\footnote{Along with the finite sort that
  constitutes the nodes of the derivation tree, a number of additional
  finite sorts and functions mapping to and from them are employed to
  represent phonetic forms, features, categories, etc.}
These axioms are derived from properties and principles of natural
language syntax that are considered \emph{universal} in so far as they
apply to all natural languages (\cite{chomsky1995minimalist,
  collins:doi:10.1111/synt.12117}).
Let us now review aspects of linguistic theory from which these axioms
are derived and discuss how these axioms constrain the model of the MG
parse tree.

In accordance with the theory of Bare Phrase Structure (BPS)
(\cite{chomsky1995minimalist}), one of the uninterpreted functions is
a binary function (over the nodes in the graph) that is constrained by
axioms that model the recursive structure building operation,
\emph{Merge} (\cite{chomsky1995minimalist,
  collins:doi:10.1111/synt.12117}); another unary functions models the
chains produced by the movement of phrases within the parse
tree.\footnote{An uninterpreted function for modeling head movement is
  also included, with relevant axioms in accordance with the
  Head-Movement constraint as given in (\cite{baker1988incorporation,
    hale1993argument, stabler2001recognizing}}
Each node in the parse tree has a \emph{head}, which is one of the
leaf nodes (which correspond to lexical items) in the parse
tree\footnote{See \cite{radford2009introduction} for a discussion of
  the \emph{Headedness Principle}, according to which \emph{``every
    nonterminal node in a syntactic structure is a projection of a
    head word.''}}; this mapping is established by a unary
uninterpreted function.
Each node in the parse tree is labeled with a category.
Categories are interpretable properties of lexical items that can
project -- the category associated with a given phrase is the category
associated with the head of that phrase.
An additional finite sort encodes the universal functional and lexical
categories $\{ C_{Declarative} , C_{Question} , T , v , V \}$ and
$\{ P , D , N\}$ (\cite{adger2011features}) and additional functions
and axioms encode the two extended projections C-T-V and P-D-N
(\cite{grimshaw2005words, adger2003core}) that constrain what
structural configurations the functional categories may be arranged in
within a derivation that converges.

Finally, we consider the axioms imposed on a given sentence by:
(a) the linear ordering of the words in a sentence --
\emph{linearization} is modeled by a conjunction of axioms for
\emph{lifting the derived tree from the derivation tree} as presented
in \cite{graf2013local}\footnote{See also
  \cite{collins:doi:10.1111/synt.12117}.} and Kayne's Linear
Correspondence Axioms (\cite{kayne1994antisymmetry});
(b) the syntactic relations annotating a sentence -- these are each
translated into axioms for either morphological agreement or
predicate-argument structure, the latter in accordance with the theory
of argument structure in (\cite{hale1993argument}).\footnote{This
  theory requires that the lexical heads associated with a predicate
  and its arguments must enter into particular structural
  configurations within a derivation. (See also
  \cite{hale2002prolegomenon}.}

\begin{table*}[t!]
  \centering
  {\small
  \begin{tabular}{lll}
    \hline
    ${I}_{i}$ & Sentence & Syntactic Relations \\
    \hline
    ${I}_{1}$ & ``John has eaten pizza.''  & agree(John, has), arg(John, eaten), arg(pizza, eaten) \\
    ${I}_{2}$ & ``Has Sally eaten pizza?''  & agree(Sally, has), arg(Sally, eaten), arg(pizza, eaten) \\
    ${I}_{3}$ & ``What has John eaten?''  & agree(John, has), arg(John, eaten), arg(What, eaten) \\
    ${I}_{4}$ & ``Who has eaten pizza?''  & agree(Who, has), arg(Who, eaten), arg(pizza, eaten) \\
    ${I}_{5}$ & ``Sally was eating pizza.''  & agree(Sally, was), arg(Sally, eating), arg(pizza, eating) \\
    ${I}_{6}$ & ``Was John eating pizza?''  & agree(John, was), arg(John, eating), arg(pizza, eating) \\
    ${I}_{7}$ & ``What was Sally eating?''  & agree(Sally, was), arg(Sally, eating), arg(What, eating) \\
    ${I}_{8}$ & ``Who was eating pizza?''  & agree(Who, was), arg(Who, eating), arg(pizza, eating) \\
    ${I}_{9}$ & ``Pizza was eaten.''  & agree(pizza, was), arg(pizza, eaten) \\
    ${I}_{10}$ & ``Was pizza eaten?''  & agree(pizza, was), arg(pizza, eaten) \\
    ${I}_{11}$ & ``What was eaten?''  & agree(What, was), arg(What, eaten) \\
    \hline
  \end{tabular}
  }
  \caption{Input sequence of annotated sentences. %
    The annotation of each sentence includes the syntactic relations
    listed for morphological agreement (indicated by \emph{agree}) and
    predicate-argument structure (indicated by \emph{arg}); 
    the type of the sentence -- i.e. either declarative or
    interrogative -- is also annotated on each sentence (but not listed here).
    The sentences listed here include passive constructions
    ($I_9, I_{10}, I_{11}$), yes/no-questions ($I_{2}, I_{6}, I_{10}$)
    and wh-questions ($I_3, I_4, I_7, I_8, I_{11}$).}
  \label{table:input}
\end{table*}

\section{Experiment}
\label{sec:experiment}
We used our implementation of the procedure presented in
\sectionref{sec:inferenceprocedure} to infer a set of minimalist
lexicons, denoted here as $G^{*}$, from an input sequence with eleven
sentences (listed in Table-\ref{table:input}), each annotated with
predicate-argument relations as well as morphological
agreement.\footnote{We bounded the acquisition model with the
  following parameters: a parse may have up to 3 instances of phrasal
  movement and up to one instance of head movement; lexical items may
  have at most 3 features.}
We validated the lexicons sampled from $G^{*}$ by using an
agenda-based MG parser (\cite{harkema2001parsing}) to verify that the
lexicon can be used to parse each sentence in the input sequence.

Manual inspection of lexicons sampled from $G^{*}$ revealed that many
of them had a large number of lexical items and produced parses
that do not resemble those found in contemporary theories of syntax --
see Lexicon-A in Figure-\ref{fig:inferredlexicons} for an example.

\begin{figure}[h!]
  \centering
  {\small
    \begin{tabular}{lll}
      Lexicon-A & Lexicon-B & Lexicon-C \\
      \hline
      eaten::\textless=x2,+l,$\sim$x1 & eaten::=x2,$\sim$x2 & eaten::=x3,$\sim$x4 \\
      eaten::=x2,+l,$\sim$x2 & eaten::=x2,+r,$\sim$x2 & eaten::=x0,=x0,$\sim$x1 \\
      eating::\textless=x2,+l,$\sim$x1 & eating::=x2,$\sim$x2 & eating::=x0,=x0,$\sim$x4 \\
      eating::=x2,+l,$\sim$x2 & has::=x1,+l,$\sim$x2 & has::=x1,+l,$\sim$x2 \\
      has::$\sim$x0,-l & has::=x2,+l,$\sim$x2 & john::$\sim$x0,-l \\
      has::=x2,+r,$\sim$x1 & has::$\sim$x2,-r,-r & pizza::$\sim$x0 \\
      john::=x0,$\sim$x2 & john::=x2,+r,$\sim$x2 & pizza::$\sim$x3,-l \\
      john::$\sim$x2,-l,-r & pizza::=x2,+r,$\sim$x2 & sally::$\sim$x0,-l \\
      pizza::=x2,$\sim$x2 & sally::=x2,+r,$\sim$x2 & was::=x4,+l,$\sim$x2 \\
      pizza::=x2,+l,$\sim$x2 & was::=x2,+l,$\sim$x2 & what::$\sim$x0,-r \\
      sally::$\sim$x2,-l,-r & was::=x2,+l,$\sim$x2 & what::$\sim$x3,-l,-r \\
      sally::=x2,+l,$\sim$x2 & what::$\sim$x2,-r,-l & who::$\sim$x0,-l,-r \\
      was::$\sim$x2,-l,-l & who::=x2,+r,$\sim$x0 & $\epsilon_{Cdecl}$::=x2,C \\
      was::=x2,+r,$\sim$x1 & $\epsilon_{Cdecl}$::\textless=x2,C & $\epsilon_{Cintr}$::\textless=x2,C \\
      what::=x1,$\sim$x1 & $\epsilon_{Cintr}$::=x2,C & $\epsilon_{Cintr}$::\textless=x2,+r,C \\
      what::=x2,+l,$\sim$x1 & $\epsilon_{Cintr}$::\textless=x0,C & ~ \\
      who::$\sim$x2,-l,-r & ~ \\
      $\epsilon_{Cdecl}$::\textless=x2,C & ~ \\
      $\epsilon_{Cdecl}$::=x1,C & ~ \\
      $\epsilon_{Cintr}$::=x1,C & ~ \\
      $\epsilon_{Cintr}$::\textless=x1,C & ~ \\
      \hline
    \end{tabular}
  }
    \caption{Examples of inferred lexicons that satisfy the conditions
      imposed by the input sequence in Table-\ref{table:input}. Each
      lexical item consists of a pairing of a phonetic form and a
      sequence of syntactic features, separated by a double-colon. The
      phonetic forms $\epsilon_{Cdecl}$ and $\epsilon_{Cintr}$ are covert
      (unpronounced); their presence in a minimalist parse tree
      indicates whether the parse is of a declarative or interrogative
      sentence, respectively.}
  \label{fig:inferredlexicons}
\end{figure}

\begin{figure}[h!]
  \centering
  \includegraphics[width=0.58\textwidth]{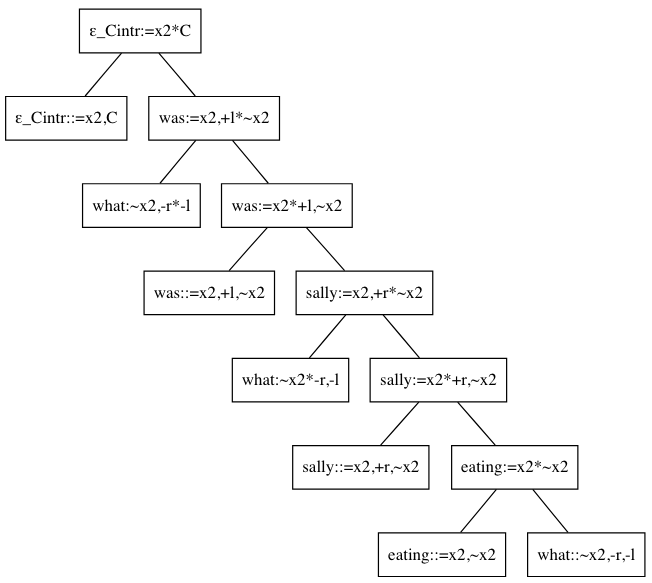}
  \caption{An MG parse for the sentence ``What was Sally eating?''
    derived from Lexicon-B in Figure-\ref{fig:inferredlexicons}.
    Although this parse satisfies the constraints imposed by the
    annotations listed for $I_7$ in Table-\ref{table:input}, it does
    not accord with the parse prescribed by contemporary theories of
    syntax -- e.g. nowhere in the parse does the auxiliary verb
    ``was'' establish a local relation with the main predicate
    ``eating.''}
  \label{fig:derivationB}
\end{figure}

We filtered out lexicons such as these by using Z3 to identify
lexicons in $G^{*}$ that were \emph{optimal} with respect to a cost
function that penalizes a lexicon for the number of lexical entries it
has.\footnote{We did this by encoding this cost function as a logical
  formula, adding it to the SMT-solver after running the inference
  procedure, and then re-solving; the resulting set of (inferred)
  minimalist grammars are optimal with respect to the specified cost
  functions.}
This produced a subset of $G^{*}$ in which every lexicon had exactly
15 lexical items, the minimal number of lexical items required for a
lexicon to be able to produce parses that accord with the specified
input sequence.
See Lexicon-B in Figure-\ref{fig:inferredlexicons} for an example of a
lexicon in this subset; see Figure-\ref{fig:derivationB} for an
example of a parse produced by Lexicon-B.

We manually inspected this subset of $G^{*}$ and found that most of
the lexicons produced parses with many instances of internal merge
that could have been eliminated without any side-effects, and
that these parses that did not accord with contemporary
theories of syntax.

\begin{figure}[h!]
  \centering
  \includegraphics[width=0.58\textwidth]{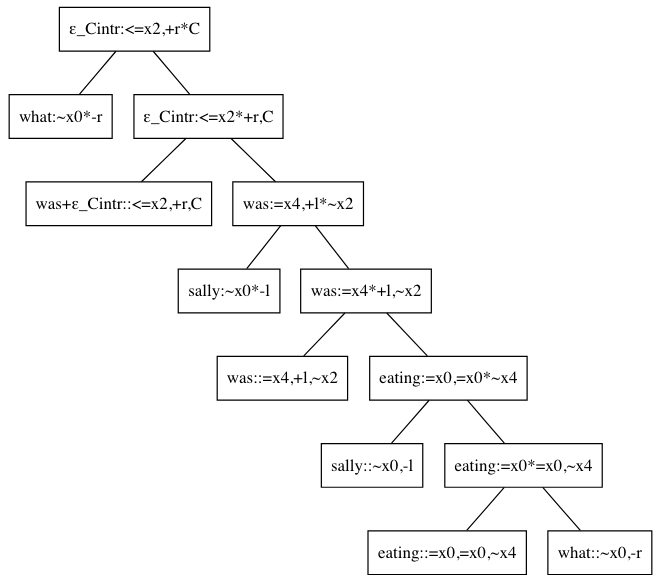}
  \caption{An MG parse for the sentence ``What was Sally eating?'' (see
    $I_7$ in Table-\ref{table:input} for annotations) derived from
    Lexicon-C in Figure-\ref{fig:inferredlexicons}.
    The parse is assembled in a bottom-up manner via merge:
    first ``eating'' merges with ``what'' and then with ``Sally'', thus
    establishing (via locality) predicate-argument relations;
    once the resulting structure then merges with ``was'', ``sally''
    then undergoes subject-raising by (internally) merging with
    ``was'', thus establishing morphological agreement between
    `''sally'' and ``was''.
    Next the head of the auxiliary verb ``was'' undergoes head
    movement to merge with the covert form, $\epsilon_{Cintr}$,
    which indicates that the sentence is an interrogative.
    Finally, ``what'' undergoes wh-fronting by (internally) merging
    with $\epsilon_{Cintr}$.
    Wh-fronting and Subject-raising are triggered by different
    licensor features, the former by $+r$ and the latter by $+l$.
    The feature sequences displayed in internal nodes have an
    asterisk separating features that have already been consumed (on
    the left) from those that have not (on the right).
    \vspace{0.05in}}
  \label{fig:derivationA}
\end{figure}

Finally, we further refined this subset of $G^{*}$ by using Z3 to
identify lexicons that were \emph{optimal} with respect to two
additional cost functions:
\begin{compactenum}
\item (minimize) the total number of selectional and licensing
  features in the lexicon and the parses; this cost function rewards
  reduction in the total size of both the lexicon and the
  derivations;\footnote{This cost function is based on the MDL
    principle (\cite{grunwald2007minimum}) as applied to MGs in
    (\cite{doi:10.1111/1467-9612.00004}).}
\item (maximize) the number of \emph{distinct} selectional features in
  the lexicon; this cost function rewards lexicons that are less
  inclusive (i.e. they are less likely to overgenerate).\footnote{This
    cost function is based on the Subset Principle
    (\cite{berwick1985acquisition}), which asserts that a language
    learner will always choose the least inclusive grammar available
    at each stage of acquisition; the adaption of this principle is a
    logical necessity if one assumes that the learner does not make
    use of (indirect) negative evidence. See also
    \cite{yang2015negative, yang2016price}. }
\end{compactenum}
\noindent This produced a subset of $G^{*}$ in which each lexicon had
exactly: 15 lexical items; 33 features in the lexicon (not including
the special feature $C$); 125 features in the parses; at least 4
distinct selectional features. See Lexicon-C in
Figure-\ref{fig:inferredlexicons} for a representative member of this
subset.
We found that the lexicons in this subset were all of the same form --
i.e. they are only differentiated by permutations of the feature
values and other symmetries in the model -- \emph{and that these
  lexicons produced parses that agreed with those prescribed by
  contemporary minimalist theories of syntax.} (as presented in
\cite{hornstein2005understanding}, \cite{adger2003core}, and
\cite{radford1997syntactic}.)
See Figure-\ref{fig:derivationA} for a parse produced by Lexicon-C
that demonstrates several of the syntactic phenomenon that Lexicon-C
models correctly (i.e. as prescribed by minimalist theories of syntax)
while respecting the syntactic relations prescribed in sentence $I_7$
of Table-\ref{table:input}.

\section{Conclusion}
\label{sec:conclusion}
In this study we have (i) proposed and implemented a procedure for
inferring MGs and (ii) used this procedure to infer an MG that closely
aligns with contemporary theories of syntax, thus demonstrating how
linguistically-relevant MGs may be identified within the inferred set
of MGs by optimizing cost functions derived from methods of inductive
inference that are relevant to cognitively-faithful models of language
acquisition. We observe that by enabling and disabling axioms in our
model, we can carry out experiments to determine which are redundant,
and \emph{thereby gain insight into whether the linguistic principles,
  from which the axioms of the system are largely derived, are
  justified or can be discarded}, thus aiding in the evaluation of the
Strong Minimalist Thesis.

Going forward, we plan to:
incorporate phase theory (\cite{chomsky2001derivation,
  chomsky2008phases}) into our model of a minimalist parse tree,
following the approach taken by \cite{chesi2007introduction};
examine the over-generations produced by the MGs inferred by our
procedure and understand how these over-generations relate to the cost
functions used by our procedure for identifying optimal grammars;
investigate the potential for this procedure to be used for producing
MG treebanks\footnote{See \cite{torr2018:P18-1055} for an alternative
  approach to developing large-scale MG treebanks.}, which may aid
treebank based parsing strategies, by extracting sets of (partially)
annotated sentences (that may be used as input to the inference
procedure) from treebanks such as PropBank
(\cite{kingsbury2002treebank}) or the UD treebanks
(\cite{nivre2016universal}).

\acks{The author would like to thank Robert C. Berwick, Sandiway Fong,
  Beracah Yankama, and Norbert Hornstein for their suggestions,
  feedback, and inspiration. Additionally, the author is very grateful
  for the financial support provided by Moody's Investor Services.}

\bibliography{learnaut-lics-2019.bib}
\end{document}